%% file: main.tex
\documentclass{article}
\usepackage[T1]{fontenc}

\usepackage{microtype}
\usepackage{graphicx}
\usepackage{subcaption}
\usepackage{booktabs}
\usepackage{array}
\usepackage{pifont}
\usepackage{multirow}
\usepackage[table]{xcolor}
\definecolor{methodblue}{rgb}{.867,.922,.969}
\usepackage{xspace}
\usepackage{xurl}
\usepackage{hyperref}

\usepackage[preprint]{icml2026}
\makeatletter
\renewcommand{\ICML@preprint}{\textit{Preprint.}}
\makeatother    
\hypersetup{pdfsubject={Self-distillation for reasoning models}}

\usepackage{amsmath}
\usepackage{amssymb}
\usepackage{mathtools}
\usepackage{amsthm}
\input{math_commands.tex}

\let\eqref\relax
\newcommand{\eqref}[1]{Eq.~(\ref{#1})}
\usepackage[capitalize,noabbrev]{cleveref}

\theoremstyle{plain}

\theoremstyle{definition}

\theoremstyle{remark}

\newcommand{\method}{WISD\xspace}
\newcommand{\methodfull}{Weakening-Induced On-Policy Self-Distillation\xspace}

\icmltitlerunning{Weaken to Strengthen}

\begin{document}

\twocolumn[
  \icmltitle{Weaken to Strengthen: Weakening-Induced On-Policy \\ Self-Distillation for Large Reasoning Models}

  \begin{icmlauthorlist}
    \icmlauthor{Yecheng Wu}{mit}
    \icmlauthor{Song Han}{mit,nv}
    \icmlauthor{Han Cai}{nv}
  \end{icmlauthorlist}

  \icmlaffiliation{mit}{Massachusetts Institute of Technology}
  \icmlaffiliation{nv}{NVIDIA}
  \icmlcorrespondingauthor{Yecheng Wu}{wyc557@mit.edu}

  \icmlkeywords{Self-Distillation, On-Policy Distillation, Reasoning,
    Large Language Models}

  \vskip 0.3in
]

\printAffiliationsAndNotice{}

\input{sec/0_abstract}
\input{figures/teaser}
\input{sec/1_intro}
\input{sec/2_related}
\input{sec/3_method}
\input{sec/4_exp}
\input{sec/5_analysis}
\input{sec/6_conclusion}

\clearpage
\setlength{\bibsep}{5pt plus 2pt minus 1pt}
\bibliography{references}
\bibliographystyle{icml2026}

\newpage
\appendix
\onecolumn
\raggedbottom
\setlength{\textfloatsep}{12pt plus 2pt minus 2pt}
\setlength{\floatsep}{10pt plus 2pt minus 2pt}
\input{sec/appendix}

\end{document}

%% file: math_commands.tex
\usepackage{amsmath,amsfonts,bm}

\def\eqref#1{equation~\ref{#1}}

\def\1{\bm{1}}

\DeclareMathAlphabet{\mathsfit}{\encodingdefault}{\sfdefault}{m}{sl}
\SetMathAlphabet{\mathsfit}{bold}{\encodingdefault}{\sfdefault}{bx}{n}

\newcommand{\E}{\mathbb{E}}

\newcommand{\KL}{D_{\mathrm{KL}}}



%% file: sec/0_abstract.tex
\begin{abstract}
Self-distillation offers a way to improve reasoning models when a suitable
external teacher is unavailable.
However, common approaches rely on privileged information to construct a
stronger self-teacher, limiting their use when only unlabeled prompts are
available.
In this paper, we introduce \methodfull{} (\method), which instead
constructs a weaker counterpart of the original model and uses the contrast
between the two as a training signal without requiring privileged information.
We explore two ways to construct the weaker model: low-bit quantization and
magnitude pruning.
We evaluate \method across diverse architectures and scales on mathematical and coding reasoning tasks.
Across these benchmarks, quantization and pruning improve average
accuracy over the original models by 4.0 and 5.5 percentage points,
respectively, achieving state-of-the-art performance even without privileged information.
Notably, pruning raises Qwen3.5-4B's accuracy from 57.7\% to 70.5\% on
HMMT~2025 and from 39.1\% to 56.7\% on LiveCodeBench~v6.
Quantization raises Qwen3.6-35B-A3B's accuracy from 68.5\% to 74.6\% and from
61.8\% to 69.5\% on the same benchmarks, respectively.
These results establish \method as a novel and effective self-distillation
framework for large reasoning models. Code is released at \url{https://github.com/jet-ai-projects/WISD}.
\end{abstract}

%% file: figures/teaser.tex
\begin{figure*}[t]
    \centering
    \includegraphics[width=\textwidth]{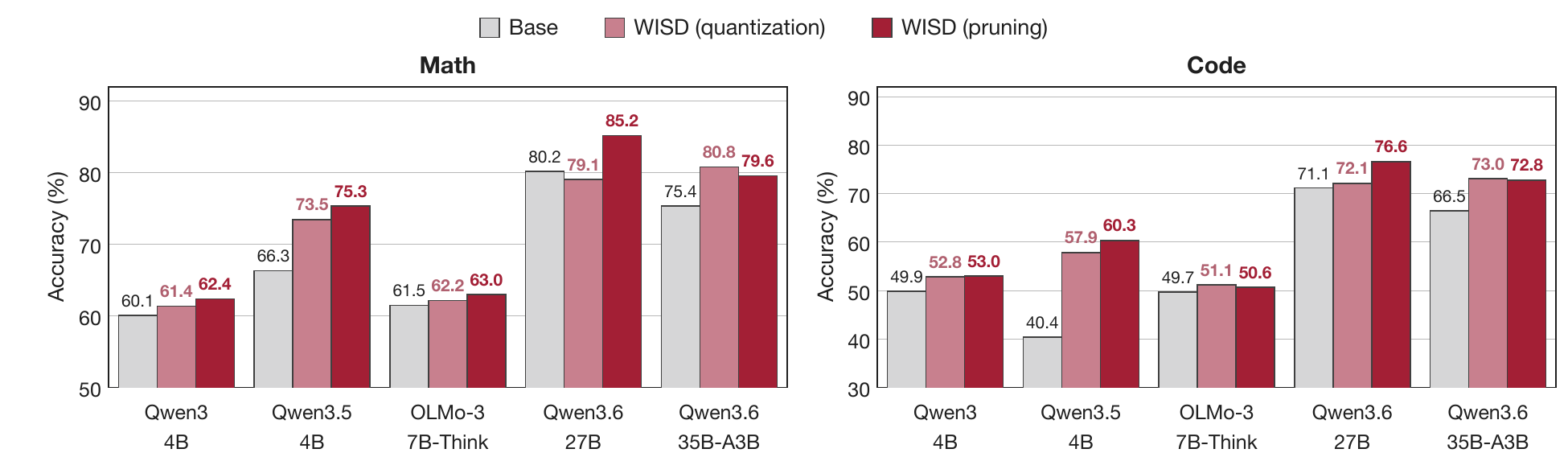}
    \caption{\textbf{Accuracy gains from self-distillation against a weakened model.}
    By constructing a weakened copy of the original model
    and using the contrast between them as a training signal, \method
    achieves consistent accuracy improvements across diverse reasoning
    models and benchmarks.
    Full results are reported in Table~\ref{tab:main-results}.}
    \label{fig:teaser}
\end{figure*}

%% file: sec/1_intro.tex
\section{Introduction}
\label{sec:intro}

On-policy distillation (OPD) has emerged as an effective approach to
improving the reasoning capabilities of large language
models~\citep{agarwal2024policy,gu2024minillm,lu2025onpolicydistillation}.
It trains a student on its own generated trajectories using dense
token-level supervision from a teacher, combining learning on the student's
own states with fine-grained feedback.
For reasoning tasks, this allows learning to address errors in
intermediate computations and decisions along the student's own solution
paths.
This approach has demonstrated substantial gains in training efficiency and
is now used in the post-training of Qwen3, MiMo-V2-Flash,
Nemotron-Cascade~2, and DeepSeek-V4~\citep{yang2025qwen3,xiao2026mimo,yang2026nemotron,xu2026deepseek}.
Recent work extends OPD to self-distillation, in which the same model serves
as both teacher and student, removing the need for a separate teacher
model~\citep{zhao2026self,shenfeld2026self,hubotter2026reinforcement}.

Self-distillation is particularly appealing when a suitable external teacher
is unavailable, especially when the goal is to improve an already capable
model~\citep{hubotter2026reinforcement}.
A common approach uses the model's in-context learning ability to construct
a stronger teacher by providing privileged information, such as reference
solutions~\citep{zhao2026self}, expert
demonstrations~\citep{shenfeld2026self}, or textual environment
feedback~\citep{hubotter2026reinforcement}.
With this additional context, the teacher can provide more informed
predictions for trajectories that the student generates from the task input
alone.
Training matches the student's next-token distributions to the teacher's,
transferring the benefit of the richer context into a policy that operates
without privileged information at inference time.

However, these approaches retain a dependence on privileged information
during training.
Reference-based methods require solutions or expert
demonstrations~\citep{zhao2026self,shenfeld2026self}, and obtaining
high-quality solutions can be
costly~\citep{zhang2026embarrassinglysimpleselfdistillationimproves}.
Feedback-based alternatives rely on informative environment responses or
reward-verified successful rollouts~\citep{hubotter2026reinforcement}.
These requirements limit their applicability when only unlabeled prompts
are available.
Even when privileged information is available, conditioning on it can shift
the teacher's predictions away from the student's and complicate knowledge
transfer~\citep{penaloza2026privileged}.
This motivates our central question: \emph{Can a model improve its reasoning
through self-distillation without privileged information?}

Recent work on weak-to-strong OPD shows that the log-probability contrast
between paired models can provide useful supervision even when both are
weaker than the student~\citep{feng2026weak,yu2026weak}.
Differences in token preferences between paired models encode a direction
of policy improvement that can be transferred to a more capable student.
From this perspective, the privileged-information approaches above take the
model itself as the weaker member of a pair and construct a stronger
counterpart through additional
context~\citep{zhao2026self,shenfeld2026self,hubotter2026reinforcement}.
We introduce \methodfull{} (\method), which reverses this construction:
we take the original model as the stronger member and create a weaker
counterpart by degrading its weights.
The contrast between the original model and its weakened copy then
provides dense supervision on the model's own trajectories, without
privileged information.
This formulation turns model weakening into a source of supervision,
enabling self-improvement using the original model and unlabeled task
prompts alone.
Table~\ref{tab:method-comparison} compares \method with representative
reasoning post-training methods in terms of learning signals and
external supervision requirements.

\input{tables/method_comparison}

Following this idea, we explore two training-free constructions of a weaker
self: low-bit quantization and magnitude pruning.
Both operate directly on model weights and require no calibration data.
Experiments across diverse model architectures and scales demonstrate
the effectiveness of \method on mathematical and coding reasoning tasks.
Figure~\ref{fig:teaser} summarizes the gains in mathematical and coding
reasoning.
Across these benchmarks, quantization and pruning improve average
accuracy over the original models by 4.0 and 5.5 percentage points,
respectively, with positive gains in 21 and
23 of the 25 model--benchmark settings, achieving state-of-the-art
performance even without privileged information.
Notably, pruning raises Qwen3.5-4B's accuracy from 57.7\% to 70.5\% on
HMMT~2025 and from 39.1\% to 56.7\% on LiveCodeBench~v6.
Quantization raises Qwen3.6-35B-A3B's accuracy from 68.5\% to 74.6\% and from
61.8\% to 69.5\% on the same benchmarks, respectively.
These results establish \method as a novel and effective self-distillation
framework for large reasoning models, requiring minimal prerequisites.

Our contributions are as follows:
\begin{itemize}
  \item We introduce \method, a self-distillation approach that treats the original model as the stronger member of a pair and learns from its contrast with a weakened copy, improving reasoning without privileged information.
  \item We instantiate this approach with two weak-model constructions, low-bit quantization and magnitude pruning, which require no calibration data and yield dense supervision on the model's own trajectories.
  \item We validate the approach across widely used reasoning models spanning diverse architectures and scales, demonstrating broad accuracy improvements on mathematical and coding reasoning tasks.
\end{itemize}

%% file: tables/method_comparison.tex
\begin{table*}[t]
  \centering
  \caption{\textbf{Comparison of training methods for reasoning.}
  \method provides dense supervision without external teachers,
  reference solutions, or environment feedback.
  Reference solutions are externally supplied answers or worked solutions
  used as teacher context. Environment feedback includes outcome
  verification and textual responses.
  GRPO uses verifiable outcome rewards. SDPO uses textual feedback or
  reward-verified self-generated solutions.}
  \label{tab:method-comparison}
  \small
  \setlength{\tabcolsep}{2pt}
  \renewcommand{\arraystretch}{1.10}
  \newcommand{\comparisonyes}{\ding{51}}
  \newcommand{\comparisonno}{\ding{55}}
  \begin{tabular}{>{\raggedright\arraybackslash}p{\dimexpr0.275\textwidth-10\tabcolsep\relax}
      >{\centering\arraybackslash}p{0.175\textwidth}
      >{\centering\arraybackslash}p{0.16\textwidth}
      >{\centering\arraybackslash}p{0.19\textwidth}
      >{\centering\arraybackslash}p{0.20\textwidth}}
    \toprule
    Method
    & \mbox{Dense Learning Signal}
    & \mbox{No External Teacher}
    & \mbox{No Reference Solutions}
    & \mbox{No Environment Feedback} \\
    \midrule
    GRPO~\citep{shao2024deepseekmath}
    & \comparisonno & \comparisonyes & \comparisonyes & \comparisonno \\
    OPD~\citep{agarwal2024policy}
    & \comparisonyes & \comparisonno & \comparisonyes & \comparisonyes \\
    OPSD~\citep{zhao2026self}
    & \comparisonyes & \comparisonyes & \comparisonno & \comparisonyes \\
    SDPO~\citep{hubotter2026reinforcement}
    & \comparisonyes & \comparisonyes & \comparisonyes & \comparisonno \\
    \midrule
    \rowcolor{methodblue}
    \textbf{\method (ours)}
    & \comparisonyes & \comparisonyes & \comparisonyes & \comparisonyes \\
    \bottomrule
  \end{tabular}
\end{table*}

%% file: sec/2_related.tex
\section{Related Work}
\label{sec:related}

\paragraph{On-Policy Distillation.}
On-policy distillation trains a student on its own generated trajectories~\citep{li2026rethinking,fu2026revisiting,ruan2026contrastive}
with token-level teacher feedback~\citep{jin2026entropy,xu2026sg,armandpour2026unmasking,wang2026not}, reducing the distribution mismatch
between training and
inference~\citep{agarwal2024policy,gu2024minillm,lu2025onpolicydistillation,wu2026lightning,xu2025speculative}.
Direct On-Policy Distillation (DOPD) and related approaches transfer policy
shifts through log-probability contrasts between paired
models~\citep{yang2026learning,feng2026weak,wu2026lightningweave}, enabling
weak-to-strong distillation~\citep{yu2026weak}.
A growing line of work extends OPD to self-distillation by conditioning the
model itself on additional information to obtain a stronger teacher:
reference solutions~\citep{zhao2026self,ke2026respecting,shen2026purified,yang2026rlsd},
demonstrations~\citep{shenfeld2026self}, and environment feedback or verified
successful rollouts~\citep{hubotter2026reinforcement,tan2026self,yang2026ogls,he2026sdzero,sun2026sropsd}.
Compared to these prior self-distillation approaches, \method requires no
privileged information. Instead, it constructs a weaker copy of the original
model and uses the contrast between the two as the training signal.
OPSA takes a complementary approach, constructing entropy-adaptive
negative advantages without an external teacher or privileged
information~\citep{ding2026opsa}.

\paragraph{Learning from Weaker Models.}
Contrasting a model with a weaker counterpart has been explored for both
generation~\citep{shi-etal-2024-trusting,hong2023selfattention,ahn2025perturbedattention} and learning.
Contrastive decoding uses a smaller amateur to guide a stronger expert at
inference time~\citep{li-etal-2023-contrastive,obrien2023contrastivedecodingimprovesreasoning}.
Variants derive weaker predictions from the model itself~\citep{gera-etal-2023-benefits,zhang2024sled,jiang-etal-2025-hicd,zhang-etal-2025-alleviating} through early
layers in DoLa~\citep{chuang2024dola}, dropout or quantization in
DCD~\citep{phan2024distillationcontrastivedecodingimproving}, and layer
pruning in PruneCD~\citep{yu-etal-2025-prunecd}.
Related small-model contrasts support decoding-time
adaptation~\citep{mitchell2024emulator,liu2024proxytuning}.
In diffusion, classifier-free guidance contrasts conditional and
unconditional predictions~\citep{ho2022cfg}, while autoguidance uses a
smaller, less-trained counterpart~\citep{karras2024guiding}.
Training-time approaches include pruned self-competitors for visual
representation learning~\citep{pmlr-v139-jiang21a} and expert--amateur
contrasts for reasoning supervision~\citep{wang-etal-2026-lightreasoner}.
Recent distillation methods construct degraded references for flow-matching
models~\citep{lin2026dreopddegradedreferenceextrapolativeonpolicy} or reduce
expert activation in an external MoE teacher to obtain a
reference~\citep{gu2026compassopdcrossfamilyonpolicydistillation}.
Self-distillation also exploits predictions conditioned on ablated visual
inputs~\citep{liang2026visualcontrastiveselfdistillation} or self-generated
negative context~\citep{pei2026negativeselfdistillationlearningreason}.
\method constructs a weaker counterpart directly through training-free
weight quantization or magnitude pruning and distills its contrast with the
original model to improve reasoning without privileged information.

%% file: sec/3_method.tex
\section{Method}
\label{sec:method}

We first review OPD and policy-shift supervision, then revisit
self-distillation from a weak-to-strong perspective to introduce \method.
Finally, we describe two training-free ways to construct its weaker model.
Figure~\ref{fig:method-overview} contrasts \method with prior
self-distillation and summarizes its training pipeline.

\input{figures/method_overview}

\subsection{Preliminaries}
\label{sec:preliminaries}

Let $\pi_\theta$ be an autoregressive student, $x\sim\mathcal{D}$ a prompt,
and $y\sim\pi_\theta(\cdot\mid x)$ a generated response.
At position $t$, the state $s_t=(x,y_{<t})$ determines a distribution over
next-token actions $a\in\mathcal{V}$.
OPD supervises the student on the states it visits during generation.
Writing $d_{\pi_\theta}$ for this state distribution, a common objective
matches the student to a teacher $\pi_T$ through reverse
KL~\citep{agarwal2024policy,gu2024minillm}:
\begin{equation}
    \mathcal{L}_{\mathrm{OPD}}(\theta)
    = \E_{s\sim d_{\pi_\theta}}
    \left[\KL\bigl(\pi_\theta(\cdot\mid s)\,\|\,\pi_T(\cdot\mid s)\bigr)\right].
    \label{eq:opd}
\end{equation}

DOPD~\citep{feng2026weak} represents supervision through a pair of frozen
models $(\pi^+,\pi^-)$, typically a post-trained model and its checkpoint
before post-training.
Their token-level log-probability difference defines a policy shift:
\begin{equation}
    r(a\mid s)
    = \log\pi^+(a\mid s)-\log\pi^-(a\mid s).
    \label{eq:policy-shift}
\end{equation}
This contrast specifies which actions the student should favor more or less,
allowing a model pair to provide useful supervision even when neither member
is stronger than the student~\citep{feng2026weak,yu2026weak}.
We next use this perspective to revisit on-policy self-distillation.

\subsection{\methodfull}
\label{sec:wisd}

Weak-to-strong distillation suggests that a useful supervision signal can
come from the difference between two models, even when a stronger teacher
is unavailable~\citep{yu2026weak}.
The central design choice is therefore how to obtain a model pair whose
relative predictions provide a learning signal.
This paired-model perspective also offers a way to understand prior
on-policy self-distillation (OPSD).
Given a model $\pi$ and privileged information $c$, these approaches
construct a more informed self-teacher $\pi(\cdot\mid s,c)$ using reference
solutions~\citep{zhao2026self}, demonstrations~\citep{shenfeld2026self},
or environment feedback~\citep{hubotter2026reinforcement}.
Viewed as a model pair, the original policy $\pi(\cdot\mid s)$ is the weaker
member, while its context-conditioned counterpart is the stronger member.
These two roles can be written as
$\pi^-(a\mid s)=\pi(a\mid s)$ and
$\pi^+(a\mid s)=\pi(a\mid s,c)$.
The student encounters the task without $c$, and the additional context
gives the teacher a more informed view of the student's decisions.
The construction therefore depends on additional information to create
the gap from which the student learns.

\method reverses this construction: we take the original model as the
stronger member and construct a weaker counterpart from its own weights.
Let $\pi_0=\pi_{\theta_0}$ denote the original model and
$\pi_w=\pi_{\mathcal{W}(\theta_0)}$ its copy under a weakening operator
$\mathcal{W}$.
Setting $(\pi^+,\pi^-)=(\pi_0,\pi_w)$ in \eqref{eq:policy-shift} gives
\begin{equation}
    r_w(a\mid s)
    = \log\pi_0(a\mid s)-\log\pi_w(a\mid s).
    \label{eq:wisd-signal}
\end{equation}
Both models receive the same prompt and generated prefix, so the contrast
compares their token preferences under an identical context.
A token receives a positive contrast when it is more likely under the
original model and a negative contrast when it is more likely under the
weak model.
Distillation uses this signal to favor tokens with larger
original-to-weak probability ratios relative to those with smaller ratios.
Using two identical policies in \eqref{eq:policy-shift} would yield zero
contrast. Weakening creates a token-dependent signal even when the student
starts from the original model.
This supplies dense self-supervision using only the original checkpoint
and unlabeled prompts.

For training, we adopt the offline DOPD procedure of
Lightning Weave~\citep{wu2026lightningweave}, with the tilted target
and distillation objective
\begin{equation}
    q_w(a\mid s)
    = \frac{\pi_0(a\mid s)\exp\!\left(r_w(a\mid s)/\alpha\right)}
    {\sum_{a'}\pi_0(a'\mid s)\exp\!\left(r_w(a'\mid s)/\alpha\right)},
    \label{eq:wisd-target}
\end{equation}
\begin{equation}
    \mathcal{L}_{\mathrm{TT}}(\theta)
    = \alpha\,\E_{s\sim d_{\pi_0}}
    \left[\KL\bigl(\pi_\theta(\cdot\mid s)\,\|\,q_w(\cdot\mid s)\bigr)\right].
    \label{eq:wisd-training}
\end{equation}
Here $\alpha>0$ is the distillation coefficient which controls how strongly
the contrast $r_w$ tilts the target away from $\pi_0$ and scales the
objective accordingly.
The original model generates trajectories on unlabeled prompts, and the
frozen pair $(\pi_0,\pi_w)$ scores their states to provide cached supervision.
A student initialized from $\theta_0$ is then trained on this supervision.
The trajectories and both scoring models remain fixed throughout optimization.
The trained student is used alone at inference time.

\subsection{Constructing a Weaker Self}
\label{sec:weak-anchors}

We instantiate $\mathcal{W}$ with low-bit quantization and magnitude
pruning, both of which operate directly on model weights without
calibration data or training.
These operators introduce two different forms of degradation: quantization
coarsens weight values, while pruning removes low-magnitude connections.
Both retain the model's architecture and tokenizer, making their
next-token predictions directly comparable to those of the original model.
The transformations apply to attention and feed-forward projection weights,
including MoE expert projections.
Embeddings, output heads, normalization parameters, and routing parameters
are preserved.
The transformation is applied only to the frozen weak scorer. The student
retains the original weights at initialization.
For each projection matrix $W$, we divide every output row into groups
$G$ of $128$ input columns and write
$M_G=\max_{j\in G}|W_{ij}|$.

\paragraph{Low-bit quantization.}
For an integer bit width $b\geq2$, let $q_b=2^{b-1}-1$.
Symmetric round-to-nearest quantization gives
\begin{equation}
    \widehat W_{ij}
    = \Delta_G\,\operatorname{clip}
    \left(\operatorname{round}\left(\frac{W_{ij}}{\Delta_G}\right),-q_b,q_b\right),
    \label{eq:weak-quantization}
\end{equation}
where $\Delta_G=M_G/q_b$.
The group maximum adapts the step size to the local weight scale.
The quantization grid contains $2^b-1$ levels, including zero.
We dequantize the resulting weights for scoring with the original model
architecture.
Our default uses $b=3$.

\paragraph{Magnitude pruning.}
For a group-relative threshold $\tau\in(0,1)$, we remove weights at or
below $\tau M_G$ within each group, then restore the original $\ell_2$
norm of each complete output row:
\begin{equation}
    \begin{aligned}
        \widetilde W_{ij}
        &= W_{ij}\,\mathbf{1}\left[|W_{ij}|>\tau M_G\right],\\
        \widehat W_{i,:}
        &= \widetilde W_{i,:}\,
        \frac{\|W_{i,:}\|_2}{\|\widetilde W_{i,:}\|_2}.
    \end{aligned}
    \label{eq:weak-pruning}
\end{equation}
The group-relative threshold accounts for variation in weight scale across
the matrix.
Its value specifies a relative magnitude threshold. The achieved sparsity
depends on the distribution of weight magnitudes.
Pruning preserves the surviving values before rescaling, while the
whole-row normalization compensates for the norm lost through deletion.
Our default uses $\tau=1/6$.
For both operators, all-zero groups or rows remain zero, and MoE experts
are processed independently.

%% file: figures/method_overview.tex
\begin{figure*}[t]
    \centering
    \includegraphics[width=\textwidth]{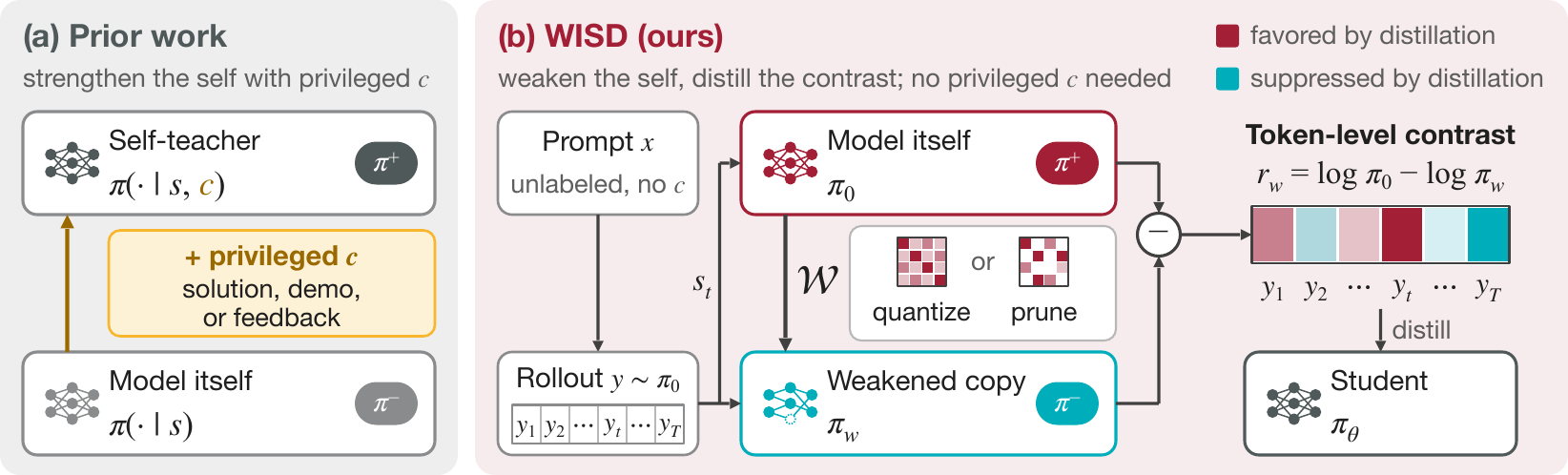}
    \caption{\textbf{Overview of \method.}
    (a) \textbf{Prior self-distillation.} Privileged information $c$, such as
    solutions, demonstrations, or feedback, strengthens the original model
    into a self-teacher.
    (b) \textbf{\method.} We instead retain the original model as the stronger
    member and construct a weaker copy through training-free quantization
    or pruning. The frozen pair scores the same self-generated trajectories,
    and its token-level log-probability contrast provides cached supervision
    for a student initialized from the original model, without privileged
    information.}
    \label{fig:method-overview}
\end{figure*}

%% file: sec/4_exp.tex
\section{Experiments}
\label{sec:experiments}

\input{tables/main_accuracy}

\subsection{Experimental Setup}
\label{sec:setup}

\paragraph{Models and weakening variants.}
We evaluate \method on five reasoning models:
Qwen3-4B~\citep{yang2025qwen3},
Qwen3.5-4B~\citep{qwen3.5},
OLMo-3-7B-Think~\citep{olmo2025olmo},
Qwen3.6-27B~\citep{qwen2026qwen36dense}, and
Qwen3.6-35B-A3B~\citep{qwen2026qwen36moe}.
These models span diverse sizes, families, and attention architectures,
covering both dense and mixture-of-experts (MoE) designs.
For each model, we compare the original checkpoint with students trained
using the quantization and pruning constructions in
Section~\ref{sec:weak-anchors}, denoted by \method{} (Quantization) and
\method{} (Pruning).
The main results use $b=3$ for quantization and $\tau=1/6$ for pruning.
Both variants initialize the student from the original model.

\paragraph{Training data.}
We train separate mathematics and coding students using 3,200 prompts
per domain from
Skywork-OR1-RL-Data~\citep{he2025skywork} and
KlearReasoner-CodeSub-15K~\citep{su2025klear}, respectively.
Each original model generates four responses per prompt to form
its own cached trajectories for self-distillation.

\paragraph{Benchmarks and metrics.}
We evaluate mathematical reasoning on
AIME~2024, AIME~2025~\citep{maa2026aime}, and
HMMT~2025~\citep{hmmt2025february}, and code generation on LiveCodeBench
(LCB) v5 and v6~\citep{jain2025livecodebench}.
We report mean accuracy over 32 samples per mathematics problem
(Avg@32) and four per coding problem (Avg@4).

\paragraph{Training and evaluation settings.}
We follow Lightning Weave's offline DOPD training
procedure~\citep{wu2026lightningweave}, using a learning rate of $10^{-6}$,
global batch size 64, and distillation coefficient $\alpha=2$.
Full training and evaluation settings are provided in
Appendix~\ref{app:details}.

\paragraph{Comparison methods.}
We include published results from OPSD~\citep{zhao2026self},
SDPO~\citep{hubotter2026reinforcement},
PW-OPSD~\citep{liu2026pwopsd},
and LOPD~\citep{zhang2026lopd} for the corresponding models and benchmarks.
These methods use reference solutions or feedback-conditioned self-teachers.

\subsection{Main Results}
\label{sec:main-results}

Table~\ref{tab:main-results} reports accuracy across five reasoning models
and five mathematics and code benchmarks.
The quantization and pruning variants of \method improve average accuracy
over the original models by 4.0 and 5.5 percentage points, respectively,
with positive gains in 21 and 23 of the 25 model--benchmark settings.
On both Qwen3-4B and OLMo-3-7B-Think, the two constructions improve the
mathematics and code averages and are competitive with OPSD and SDPO.
Both variants also surpass the reported five-benchmark averages of baselines that use privileged information.
The largest gains occur on Qwen3.5-4B, where quantization and pruning increase the five-benchmark average by 11.3 and 13.4 points, respectively.
In particular, pruning raises its HMMT~2025 accuracy from 57.7\% to 70.5\%
and LCB v6 accuracy from 39.1\% to 56.7\%.
The gains also extend to larger dense and MoE models, where
pruning improves Qwen3.6-27B's average accuracy by 5.2 points and
quantization improves Qwen3.6-35B-A3B's by 5.9 points.
On the latter, HMMT~2025 and LCB v6 accuracies rise from 68.5\% to 74.6\%
and from 61.8\% to 69.5\%, respectively.
These results demonstrate that \method turns simple weight degradation into effective
self-distillation supervision, improving reasoning accuracy across model
families, scales, and architectures without privileged information.

\subsection{Ablation Study}
\label{sec:ablation}

We study quantization precision and pruning threshold using Qwen3.5-4B
on HMMT25 (Avg@32), varying the hyperparameters $b$ and $\tau$ to control the strength of quantization and pruning respectively while keeping all other settings the same as \ref{sec:setup}.
The sweeps cover $b\in\{8,6,4,3,2\}$ and
$\tau\in\{1/16,1/8,1/6,1/4,1/2\}$.
We report both student accuracy and average generated length.

\input{figures/ablation}

\paragraph{Quantization strength.}
Figure~\ref{fig:construction-ablation}(a) shows a non-monotonic relationship
between weight precision and student accuracy.
Reducing precision from 8 to 3 bits raises accuracy from 56.04\% to 68.33\%,
while a further reduction to 2 bits lowers it to 49.90\%.
After self-distillation, the student also generates shorter responses than the Base model.
This pattern suggests that mild quantization provides too little contrast
for substantial gains over the original model, whereas excessive
quantization erodes the weak model's reasoning ability and allows noise
to overwhelm useful contrast. Moderate quantization balances contrast strength and signal quality, with $b=3$ giving the best result in this sweep and serving as our default.
\enlargethispage{6pt}

\paragraph{Pruning threshold.}
Figure~\ref{fig:construction-ablation}(b) shows a similar intermediate
range of effective constructions.
Accuracy rises from 60.94\% at $\tau=1/16$ to 69.38\% at $\tau=1/8$
and 70.52\% at the default $\tau=1/6$, then falls to 54.79\% and 49.58\%
at $\tau=1/4$ and $1/2$.
After self-distillation, the student also generates shorter responses than the Base model.
Similarly, the pruning results favor an intermediate level of damage
that creates informative contrast while preserving the weak model's
reasoning ability.
Our default threshold $\tau=1/6$ therefore combines the highest accuracy
with the shortest responses in this sweep.

%% file: tables/main_accuracy.tex
\begin{table*}[!t]
  \centering
  \caption{Accuracy (\%) across mathematics and code benchmarks. Training and evaluation settings are described in Section~\ref{sec:setup}.
  \textbf{Bold} and \underline{underlined} values denote the best and
  second-best reported results per model.
  A dash denotes an unavailable result. Math Avg.\ averages three benchmarks,
  Code Avg.\ averages two, and the final Avg.\ weights all five equally. \method achieves consistent improvements across models and benchmarks and remains competitive with other on-policy self-distillation methods without using any privileged information.}
  \label{tab:main-results}
  \small
  \setlength{\tabcolsep}{2.0pt}
  \renewcommand{\arraystretch}{1.22}
  \begin{tabular}{>{\raggedright\arraybackslash}p{0.27\textwidth}
      *{8}{>{\centering\arraybackslash}p{\dimexpr(0.73\textwidth-18\tabcolsep)/8\relax}}}
    \toprule
    \multirow{2}{*}{\textbf{Method}}
    & \multicolumn{4}{c}{\textbf{Math Reasoning}}
    & \multicolumn{3}{c}{\textbf{Code Generation}}
    & \multirow{2}{*}{\textbf{Avg.}} \\
    \cmidrule(lr){2-5}\cmidrule(lr){6-8}
    & \textbf{AIME24} & \textbf{AIME25} & \textbf{HMMT25} & \textbf{Avg.}
    & \textbf{LCB v5} & \textbf{LCB v6} & \textbf{Avg.} & \\
    \midrule
    \rowcolor{gray!12}
    \multicolumn{9}{l}{\textbf{Qwen3-4B}} \\
    Original & 73.4 & 65.6 & 41.1 & 60.07 & 53.0 & 46.8 & 49.86 & 55.98 \\
    OPSD~\citep{zhao2026self} & 75.2 & \underline{66.7} & \underline{43.9} & 61.92 & 42.7 & 35.1 & 38.88 & 52.70 \\
    SDPO~\citep{hubotter2026reinforcement} & 70.6 & 62.8 & 40.8 & 58.06 & 40.9 & 34.4 & 37.61 & 49.88 \\
    PW-OPSD~\citep{liu2026pwopsd} & \textbf{76.2} & \textbf{67.8} & 43.3 & \textbf{62.44} & --- & --- & --- & --- \\
    LOPD~\citep{zhang2026lopd} & --- & --- & --- & --- & 50.5 & 45.0 & 47.79 & --- \\
    \rowcolor{methodblue}
    \method{} (Quantization) & 75.4 & 65.8 & 42.8 & 61.35 & \textbf{56.4} & \underline{49.2} & \underline{52.80} & \underline{57.93} \\
    \rowcolor{methodblue}
    \method{} (Pruning) & \underline{75.8} & 66.0 & \textbf{45.2} & \underline{62.36} & \underline{55.7} & \textbf{50.2} & \textbf{52.96} & \textbf{58.60} \\
    \midrule
    \rowcolor{gray!12}
    \multicolumn{9}{l}{\textbf{Qwen3.5-4B}} \\
    Original & 74.3 & 67.0 & 57.7 & 66.32 & 41.7 & 39.1 & 40.39 & 55.95 \\
    \rowcolor{methodblue}
    \method{} (Quantization) & \textbf{80.2} & \underline{72.1} & \underline{68.1} & \underline{73.47} & \underline{60.8} & \underline{55.0} & \underline{57.86} & \underline{67.23} \\
    \rowcolor{methodblue}
    \method{} (Pruning) & \underline{78.8} & \textbf{76.8} & \textbf{70.5} & \textbf{75.35} & \textbf{64.0} & \textbf{56.7} & \textbf{60.33} & \textbf{69.34} \\
    \midrule
    \rowcolor{gray!12}
    \multicolumn{9}{l}{\textbf{OLMo-3-7B-Think}} \\
    Original & \underline{74.0} & 65.3 & \underline{45.2} & 61.49 & 52.1 & 47.3 & 49.71 & 56.78 \\
    OPSD~\citep{zhao2026self} & 72.8 & 67.4 & 43.9 & 61.36 & 44.1 & 45.0 & 44.57 & 54.64 \\
    SDPO~\citep{hubotter2026reinforcement} & --- & --- & --- & --- & 49.8 & 43.5 & 46.67 & --- \\
    PW-OPSD~\citep{liu2026pwopsd} & 72.6 & \textbf{69.6} & 43.8 & 62.01 & --- & --- & --- & --- \\
    LOPD~\citep{zhang2026lopd} & --- & --- & --- & --- & \textbf{53.4} & 45.8 & 49.61 & --- \\
    \rowcolor{methodblue}
    \method{} (Quantization) & \textbf{74.6} & 67.4 & 44.5 & \underline{62.15} & \underline{52.9} & \textbf{49.4} & \textbf{51.15} & \underline{57.75} \\
    \rowcolor{methodblue}
    \method{} (Pruning) & 72.7 & \underline{69.1} & \textbf{47.2} & \textbf{62.99} & 52.1 & \underline{49.2} & \underline{50.65} & \textbf{58.05} \\
    \midrule
    \rowcolor{gray!12}
    \multicolumn{9}{l}{\textbf{Qwen3.6-27B}} \\
    Original & \underline{86.8} & 78.8 & \underline{75.1} & \underline{80.21} & 73.2 & \underline{69.1} & 71.15 & \underline{76.58} \\
    \rowcolor{methodblue}
    \method{} (Quantization) & 85.1 & \underline{78.9} & 73.2 & 79.06 & \underline{76.5} & 67.7 & \underline{72.14} & 76.29 \\
    \rowcolor{methodblue}
    \method{} (Pruning) & \textbf{88.8} & \textbf{84.4} & \textbf{82.5} & \textbf{85.21} & \textbf{80.1} & \textbf{73.1} & \textbf{76.60} & \textbf{81.76} \\
    \midrule
    \rowcolor{gray!12}
    \multicolumn{9}{l}{\textbf{Qwen3.6-35B-A3B}} \\
    Original & 83.3 & 74.3 & 68.5 & 75.38 & 71.1 & 61.8 & 66.49 & 71.82 \\
    \rowcolor{methodblue}
    \method{} (Quantization) & \underline{85.9} & \textbf{81.9} & \textbf{74.6} & \textbf{80.80} & \textbf{76.6} & \textbf{69.5} & \textbf{73.04} & \textbf{77.69} \\
    \rowcolor{methodblue}
    \method{} (Pruning) & \textbf{86.3} & \underline{80.8} & \underline{71.7} & \underline{79.58} & \underline{76.3} & \underline{69.3} & \underline{72.76} & \underline{76.86} \\
    \bottomrule
  \end{tabular}
\end{table*}

%% file: figures/ablation.tex
\begin{figure*}[t]
  \centering
  \includegraphics[width=\textwidth]{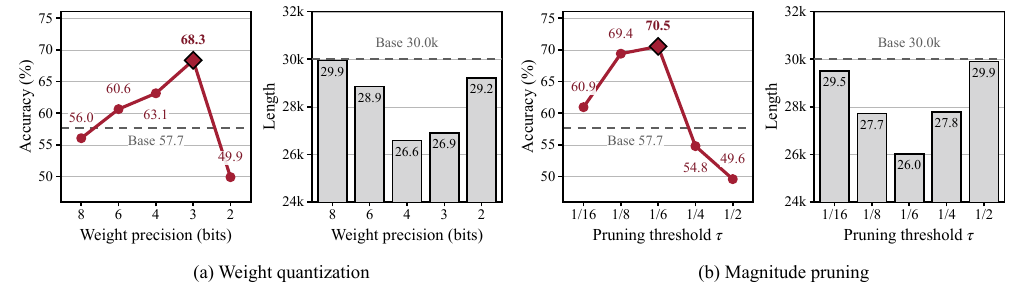}
  \caption{Ablations of weak-model construction on Qwen3.5-4B, evaluated on HMMT25.
(a) Quantization bit width. (b) Pruning threshold.
Each pair shows accuracy (Avg@32, left plot) and average generated length
in thousands of tokens (right plot). Dashed lines denote Base accuracy and
average generated length. Outlined diamonds mark the
default settings $b=3$ and $\tau=1/6$.
Both constructions achieve their highest accuracy at intermediate weakening
strengths. This pattern highlights the importance of balancing contrast
strength and signal quality for effective self-distillation.}
  \label{fig:construction-ablation}
\end{figure*}

%% file: sec/5_analysis.tex
\vspace{0pt plus 2fil}
\section{Analysis}
\label{sec:analysis}
\vspace{0pt plus 1fil}

We analyze the effectiveness of weak-model constructions and the resulting distillation signals on Qwen3-4B.
\newpage

\subsection{Effectiveness of Weak-Model Constructions}
\label{sec:construction-comparison}

\paragraph{Alternative quantization and pruning methods.}
To evaluate the effectiveness of weak-model construction in providing
contrast signals, we also explore alternative methods, including
GPTQ~\citep{frantar2023gptq} with 3-bit weights and W4A4 with and without SmoothQuant-style
smoothing~\citep{xiao2023smoothquant} for quantization,
and Wanda-based pruning~\citep{sun2024wanda} and
SparseGPT~\citep{frantar2023sparsegpt}, both with a 2:4 sparsity pattern,
for pruning.
We train separate students for mathematics and code using the identical training settings in Section~\ref{sec:setup}.

\input{tables/construction_comparison}

Table~\ref{tab:construction-comparison} shows that Smooth W4A4,
Wanda, and SparseGPT provide useful training signals on both tasks,
whereas GPTQ and plain W4A4 do not yield consistent gains.
The different outcomes of GPTQ and RTN at 3-bit precision, and of
W4A4 and Smooth W4A4 at 4-bit precision, show that bit width alone
does not determine the training value of a weak model.
The gains obtained with multiple quantization and pruning methods
further support the generality of weak-model construction as a source
of self-distillation signals.
Additional results are presented in
Appendix~\ref{app:construction-comparison}.

\phantomsection
\paragraph{Confidence-based simple baselines.}
\label{sec:confidence-controls}
To further assess the role of weight weakening, we consider two
alternative training signals derived from the original model's predictions.
\emph{Self-enhancing} directly uses the frozen original model's
token log-probabilities as rewards.
\emph{Temperature weak} softens the original model's output distribution
with a temperature of $T=2$, and uses the log-probability difference
between the original and softened distributions as the reward.
Implementation details are provided in
Appendix~\ref{app:confidence-controls}. Table~\ref{tab:construction-comparison} shows the two
confidence-based alternatives remain close to the original model.
These results show that our quantization and pruning constructions provide
effective, structured contrast signals for self-distillation, going beyond simply reinforcing the original model's confidence and leading to improved reasoning capabilities.

\input{tables/token_categories}
\input{figures/case_study}

\subsection{Understanding the Training Signal}
\label{sec:preference-analysis}

We analyze how weak-model contrasts allocate supervision across reasoning
roles on 64 Qwen3-4B mathematics training trajectories.
We label response segments using the ThinkARM taxonomy~\citep{li2026thinkarm}
and measure each role's share of the initial logit-gradient L1 magnitude.
Table~\ref{tab:token-categories} shows that quantization and pruning allocate
about 22\% of the signal to exploration and 11\% to verification, exceeding
these roles' token shares of 14.8\% and 8.1\%.
Both constructions thus place greater training weight on exploring
solution paths and checking intermediate reasoning.
Appendix~\ref{app:signal-analysis} provides the analysis details.

Figure~\ref{fig:case-study} illustrates how this supervision acts on tokens
that support intermediate calculations and logical checks.
In the paint-coverage example, quantization increases the target probabilities
of a digit in the coverage calculation and the negation \texttt{not},
reinforcing the conclusion that the proposed amount of paint is insufficient.
In the algebra example, pruning increases the target probability of
\texttt{0} in an intermediate result, supporting the exclusion of a candidate
root that makes the original divisor zero.
These cases show how the contrast reinforces tokens that express numerical
results and their logical consequences, supplying supervision for the local
decisions that support correct reasoning.

%% file: tables/construction_comparison.tex
\begin{table}[!ht]
  \centering
  \caption{Qwen3-4B student accuracy (\%) with weak-model constructions
  and simple baselines on AIME24 (Avg@32) and LCB v5 (Avg@4).
  Multiple quantization and pruning constructions improve accuracy on both tasks.
  Our default RTN and magnitude-pruning constructions outperform both simple
  baselines, supporting the value of structured contrast beyond confidence
  reinforcement. Bold marks the highest displayed accuracy within the
  quantization and pruning families.}
  \label{tab:construction-comparison}
  \small
  \setlength{\tabcolsep}{3pt}
  \begin{tabular}{
      >{\raggedright\arraybackslash}p{\dimexpr0.64\columnwidth-2\tabcolsep\relax}
      >{\raggedleft\arraybackslash}p{\dimexpr0.18\columnwidth-2\tabcolsep\relax}
      >{\raggedleft\arraybackslash}p{\dimexpr0.18\columnwidth-2\tabcolsep\relax}}
    \toprule
    Method & AIME24 & LCB v5 \\
    \midrule
    Qwen3-4B & 73.4 & 53.0 \\
    \midrule
    \multicolumn{3}{l}{\textit{Quantization}} \\
    \addlinespace[2pt]
    GPTQ (3-bit)~\citep{frantar2023gptq} & 72.0 & 53.4 \\
    W4A4 & 70.0 & 53.0 \\
    Smooth W4A4~\citep{xiao2023smoothquant} & 75.3 & 55.2 \\
    \rowcolor{methodblue}
    RTN (3-bit) & \textbf{75.4} & \textbf{56.4} \\
    \midrule
    \multicolumn{3}{l}{\textit{Pruning}} \\
    \addlinespace[2pt]
    Wanda~\citep{sun2024wanda} & 74.4 & 55.6 \\
    SparseGPT~\citep{frantar2023sparsegpt} & 73.5 & \textbf{55.7} \\
    \rowcolor{methodblue}
    Magnitude pruning & \textbf{75.8} & \textbf{55.7} \\
    \midrule
    \multicolumn{3}{l}{\textit{Simple baselines}} \\
    \addlinespace[2pt]
    Self-enhancing & 72.8 & 53.6 \\
    Temperature weak ($T=2$) & 73.1 & 53.0 \\
    \bottomrule
  \end{tabular}
\end{table}

%% file: tables/token_categories.tex
\begin{table*}[!t]
  \centering
  \caption{Distribution of initial training signals across reasoning roles
  (\%). The first row gives token frequencies. Remaining rows give shares
  of logit-gradient L1 magnitude. Bold marks shares above token frequency.
  Quantization and pruning allocate about 22\% of the signal to exploration
  and 11\% to verification, exceeding these roles' token shares of 14.8\%
  and 8.1\%.
  Both constructions thus place greater training weight on exploring
  solution paths and checking intermediate reasoning.}
  \label{tab:token-categories}
  \normalsize
  \setlength{\tabcolsep}{6pt}
  \begin{tabular}{lrrrrrrrrr}
    \toprule
    Method & Read & Analyze & Plan & Implement & Explore & Verify
      & Monitor & Answer & Formatting \\
    \midrule
    All tokens & 9.1 & 24.3 & 6.6 & 28.2 & 14.8 & 8.1 & 1.4 & 1.8 & 5.6 \\
    \midrule
    \rowcolor{methodblue}
    Quantization & 7.6 & \textbf{27.1} & \textbf{8.4} & 16.0 & \textbf{22.3} & \textbf{11.2} & \textbf{2.4} & \textbf{2.0} & 3.0 \\
    \rowcolor{methodblue}
    Pruning & 7.8 & \textbf{26.7} & \textbf{8.3} & 15.9 & \textbf{22.1} & \textbf{11.4} & \textbf{2.5} & \textbf{2.1} & 3.2 \\
    \bottomrule
  \end{tabular}
\end{table*}

%% file: figures/case_study.tex
\setlength{\dbltextfloatsep}{8pt plus 2pt minus 2pt}
\setlength{\dblfloatsep}{6pt plus 2pt minus 2pt}
\begin{figure*}[!t]
  \centering
  \includegraphics[width=\textwidth]{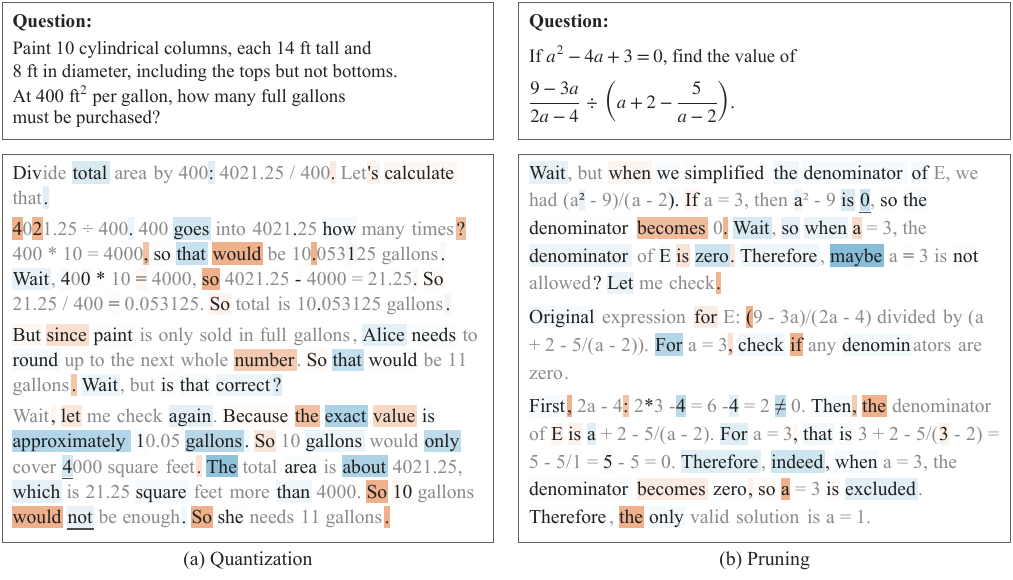}
  \caption{\textbf{Token-level effects of \method.}
  Original-model trajectories illustrate a coverage check with
  quantization (a) and a zero-divisor check with pruning (b).
  Blue and orange indicate increased and decreased target probability
  relative to the behavior distribution.
  Quantization reinforces tokens supporting the coverage calculation and the
  conclusion that ten gallons are insufficient, while pruning reinforces
  the intermediate zero used to exclude an invalid root.
  These examples illustrate how weak-model contrasts provide local
  supervision for numerical checks and the logical decisions they support.}
  \label{fig:case-study}
\end{figure*}

%% file: sec/6_conclusion.tex
\section{Conclusion}
\label{sec:conclusion}

We introduced \method, a self-distillation framework that improves reasoning
models without privileged information.
The core idea is to construct a weaker counterpart of the original model
and turn their token-level log-probability contrast into dense supervision
on self-generated trajectories.
We instantiate this idea with low-bit quantization and magnitude pruning,
enabling training from the original checkpoint and unlabeled prompts.
Experiments across diverse reasoning models and benchmarks demonstrate
broad accuracy gains. Our analyses show that weak-model contrasts
emphasize exploration and verification while providing training directions
beyond confidence reinforcement.
These results establish model weakening as an effective source of
supervision for improving reasoning through self-distillation.

%% file: sec/appendix.tex
\section{Experimental Details}
\label{app:details}

\subsection{Training Data and Optimization}

We train separate mathematics and code students using 3,200 prompts from
Skywork-OR1-RL-Data~\citep{he2025skywork} and
KlearReasoner-CodeSub-15K~\citep{su2025klear}, respectively.
Each original model samples four responses per prompt, producing 12,800
cached trajectories per domain.
Students initialize from their original checkpoints and use Lightning
Weave's offline DOPD procedure~\citep{wu2026lightningweave} with the frozen
original-to-weak contrast.
Table~\ref{tab:training-settings} summarizes the shared configuration.

\input{tables/training_settings}

\subsection{Evaluation and Published Baselines}

\paragraph{Benchmarks and metrics.}
AIME~2024, AIME~2025, and HMMT~2025 each contain 30 problems, evaluated
with 32 responses per problem.
LiveCodeBench v5 and v6 contain 279 and 131 problems, respectively,
evaluated with four responses per problem.
Avg@$n$ is the fraction of correct responses across all problems and
their $n$ samples.
Math Avg.\ equally weights the three mathematics benchmarks, Code Avg.\
equally weights the two coding benchmarks, and the final Avg.\ equally
weights all five.
We compute averages from unrounded scores.
AIME~2024 also serves as the development benchmark for exploring
weak-model constructions.

\paragraph{Generation settings.}
We use each model's native chat template with thinking enabled,
temperature $0.6$, top-$p=0.95$, and top-$k$ disabled.
Our main-table evaluations use a 40,960-token output limit and a 49,152-token
context window. Mean response length counts generated tokens, excluding the prompt.

\paragraph{Published results.}
OPSD and PW-OPSD mathematics results come from Table~4 of
\citet{liu2026pwopsd}, with the Moderate schedule for PW-OPSD.
These use Avg@12 averaged over three evaluation seeds, temperature $1.0$,
top-$p=0.95$, and a 38,912-token output limit.
Qwen3-4B SDPO mathematics results come from Table~1 of
\citet{wu2026dapddualanchoredpolicydistillation}, also using Avg@12,
temperature $1.0$, top-$p=0.95$, and a 38,912-token output limit.
OPSD, SDPO, and LOPD code results come from Table~2 of
\citet{zhang2026lopd}, using pass@1 on the same LCB tasks,
temperature $0.7$, top-$p=0.95$, top-$k=20$, and a 16,384-token output
limit.

\input{sec/appendix_eval_templates}

\subsection{Weak-Model Construction Details}
\label{app:construction-settings}

\paragraph{Quantization.}
RTN uses symmetric group-wise quantization with 128 input coordinates
per output-row group, except for the group-32 variant.
The signed grid has $2^b-1$ levels, so 2-bit RTN uses
$\{-1,0,1\}$.
W4A8 and W4A4 combine 4-bit RTN weights with per-token activation
quantization at 8 and 4 bits, respectively.
Smooth W4A4 applies SmoothQuant-style input-channel and weight rescaling
with smoothing exponent $0.5$ before quantization~\citep{xiao2023smoothquant}.
GPTQ uses 3-bit or 2-bit weights, group size 128, and damping
$0.01$~\citep{frantar2023gptq}.

\paragraph{Pruning.}
Our default applies the group-relative threshold $\tau=1/6$ and restores
each complete output row's $\ell_2$ norm, as defined in
Section~\ref{sec:weak-anchors}.
The additional per-row magnitude variants remove the smallest 20\% or
40\% of weights in each row.
Wanda ranks weights by magnitude times the clean-model input-activation
norm, using either per-row sparsity or a 2:4
pattern~\citep{sun2024wanda}.
SparseGPT uses input covariance statistics with damping $0.01$, at
50\% sparsity or with a 2:4 pattern~\citep{frantar2023sparsegpt}.
MLP channel pruning removes the lowest-ranked 25\% of channels according
to the product of the down-projection column norm and its input-activation
norm, jointly masking the corresponding gate/up rows and down columns.

\paragraph{Calibration and scoring.}
The calibrated alternatives use 128 cached training sequences from the
corresponding domain, with up to 2,048 tokens per sequence.
Calibration statistics are collected from the unchanged original model.
GPTQ and SparseGPT accumulate and factorize their covariance matrices in FP64.
The Qwen3-4B construction studies use FP32 scorer forwards and BF16 student
training.
The Qwen3.5-4B sweep uses BF16 scorers, except for the default pruning
student, which uses the FP32-scored run from the main comparison.
All constructions modify the frozen weak scorer and retain the student's
original architecture.

\subsection{Simple Baselines}
\label{app:confidence-controls}

We use two additional signals derived from the original model for the
comparison in Table~\ref{tab:construction-comparison}.
\emph{Self-enhancing} uses its token log-probability directly:
\begin{equation}
  r_{\mathrm{self}}(a\mid s)=\log\pi_0(a\mid s).
\end{equation}
\emph{Temperature weak} uses the contrast with a softened output distribution:
\begin{equation}
  r_T(a\mid s)=\log\pi_0(a\mid s)-\log\pi_{0,T}(a\mid s),
  \qquad
  \pi_{0,T}(\cdot\mid s)=\operatorname{softmax}(z_0(s)/T),
  \quad T=2.
\end{equation}
Here $z_0$ denotes the original model's logits.
Both signals use full-vocabulary-normalized probabilities on the same
cached token candidates and the training configuration in
Table~\ref{tab:training-settings}.
Both assign zero reward to \textsc{Other}.
We train separate mathematics and code students from Qwen3-4B and
evaluate their final checkpoints on AIME24 and LCB v5.
AIME24 and LCB v5 use a 40,960-token output limit and 49,152-token context.

\input{sec/appendix_analysis}

%% file: tables/training_settings.tex
\begin{table}[!htbp]
  \centering
  \caption{Training configuration for \method.
  Mathematics and code students use separate prompt sets and cached trajectories.}
  \label{tab:training-settings}
  \small
  \setlength{\tabcolsep}{6pt}
  \begin{tabular*}{\textwidth}{@{\extracolsep{\fill}}llll@{}}
    \toprule
    Rollout parameter & Value & Training parameter & Value \\
    \midrule
    Prompts per domain & 3,200 & Distillation coefficient $\alpha$ & 2 \\
    Responses per prompt & 4 & Learning rate & $10^{-6}$, constant \\
    Trajectories per domain & 12,800 & Global batch size & 64 \\
    Temperature / top-$p$ & 1.0 / 1.0 & Optimizer updates & 200 \\
    Default response limit & 2,048 tokens & Student precision & BF16 \\
    Rollout seed & 42 & Training seed & 1234 \\
    \bottomrule
  \end{tabular*}
\end{table}

%% file: sec/appendix_eval_templates.tex
\subsection{Evaluation Prompts and Answer Extraction}
\label{app:evaluation-prompts}

\paragraph{Mathematics prompt.}
Before application of the model's native chat template, each mathematics
example is represented exactly as
\begin{quote}
\small\ttfamily\raggedright
Question: \{Problem\}\\
Answer:
\end{quote}
We enable the model's native reasoning mode where available and stop on its
end-of-sequence or end-of-turn token.  For every problem we independently
sample 64 responses at temperature 0.6 and top-$p=0.95$, allowing at most
32,768 generated tokens.  We extract the last balanced
\texttt{\textbackslash boxed\{...\}} expression, remove superficial
formatting such as whitespace, delimiters, and units, and canonicalize integer
answers before exact comparison with the reference.  We do not use a
computer-algebra system to grant additional symbolic equivalences.  A
length-truncated response is still scored if it contains a valid extracted
answer.

\paragraph{LiveCodeBench prompt.}
We use LiveCodeBench's \texttt{CodeQwenInstruct} generation format.  Its user
message begins with:
\begin{quote}
\small\ttfamily\raggedright
You will be given a question (problem specification) and will generate a
correct Python program that matches the specification and passes all tests.
You will NOT return anything except for the program.\\[2pt]
Question: \{question\_content\}
\end{quote}
For problems with starter code, the message then asks the model to complete
the supplied code and enclose the answer in code delimiters.  Otherwise it
instructs the model to read from standard input, write to standard output, and
return the program inside a Python code block initialized with
\texttt{\# YOUR CODE HERE}.  The official system wrapper is
\texttt{You are a helpful assistant.}, expressed with the Qwen instruction
delimiters.  We sample four responses per problem at temperature 0.6 and
top-$p=0.95$, with a maximum of 40,960 generated tokens.

We use LiveCodeBench's official code extraction and test execution.  A sample
is correct only if its extracted program passes every test for that problem.
Tests run in isolated, read-only containers with the benchmark's fast-test
mode and a six-second timeout per test.  Runtime errors, timeouts, extraction
failures, and incomplete programs are scored as incorrect.

%% file: sec/appendix_analysis.tex
\section{Additional Experimental Results}
\label{app:ablation-analysis}

\subsection{Alternative Weak-Model Constructions}
\label{app:construction-comparison}

Table~\ref{tab:construction-comparison-full} extends the construction
comparison in Section~\ref{sec:construction-comparison} with additional
precisions, group sizes, sparsity patterns, and MLP channel pruning.
All students initialize from Qwen3-4B, use the configuration in
Table~\ref{tab:training-settings}, and are evaluated on AIME24 with Avg@32.

\input{tables/construction_comparison_full}

Smooth W4A4 improves on plain W4A4 at the same weight and activation
precision.
Per-row magnitude pruning and Wanda both improve accuracy at 40\%
sparsity, while the default group-relative magnitude rule gives the
strongest pruning result.

\subsection{Row-Norm Restoration}
\label{app:row-norm-restoration}

We compare magnitude pruning with and without the row-norm restoration
in \eqref{eq:weak-pruning}.
Both constructions use group size 128 and $\tau=1/6$, with the same
pruning mask, cached trajectories, student initialization, and training
recipe.
The paired scorers use FP32 forwards.
Table~\ref{tab:row-norm-restoration} reports student accuracy on Qwen3-4B
with AIME25 and Qwen3.5-4B with HMMT25, using Avg@32, a 40,960-token
output limit, and a 49,152-token context.

\input{tables/row_norm_restoration}

Row-norm restoration adds 0.6 and 0.5 percentage points on the two models.
We retain it in our default construction, following prior work on
rescaling retained weights after pruning~\citep{hayou2021robust}.
Restoring each row's original $\ell_2$ norm preserves its weight scale
while retaining the directional change introduced by pruning.

\section{Training-Signal Analysis}
\label{app:signal-analysis}

\subsection{Token-Level Training Signal}

At each state $s$, the cache retains 16 original-model token candidates
$\mathcal{S}$ and aggregates the remaining vocabulary into an
\textsc{Other} bucket.
Let $b$ be the cached behavior distribution over these 17 buckets and
$m=\sum_{j\in\mathcal{S}}b_j$.
Scorer probabilities are normalized over the full vocabulary before
gathering the candidates.
We set $r_j=r_w(a_j\mid s)$ for retained candidates and
$r_{\textsc{Other}}=0$.

The offline implementation uses the tilted target with
behavior-weighted log-probability regression:
\begin{equation}
  q_j=\frac{b_j\exp(r_j/\alpha)}
  {\sum_k b_k\exp(r_k/\alpha)},
  \qquad
  \ell(p)=\frac{\alpha}{2m}\sum_j b_j
  \bigl(\log p_j-\log q_j\bigr)^2,
  \label{eq:appendix-target-loss}
\end{equation}
where $p=\operatorname{softmax}(z)$ is the student's bucket distribution.
At $p=b$, the initial gradient with respect to the bucket logits is
\begin{equation}
  g_j=\left.\frac{\partial\ell}{\partial z_j}\right|_{p=b}
  =-\frac{b_j}{m}\left(r_j-\sum_k b_k r_k\right).
  \label{eq:initial-gradient}
\end{equation}
This expression centers each candidate's contrast against the
behavior-weighted mean.
For each position $t$, we measure the signal magnitude by
$s_t=\|g_t\|_1$.
The share assigned to a reasoning role $C$ is
\begin{equation}
  \operatorname{SignalShare}(C)
  =100\,\frac{\sum_{t\in C}s_t}{\sum_t s_t}.
  \label{eq:role-signal-share}
\end{equation}
The token-share baseline is the fraction of positions assigned to that
role.
Quantization and pruning are evaluated on the same prefixes, token
candidates, and behavior probabilities.

\subsection{Contextual Reasoning Roles}
\label{app:reasoning-roles}

We select every 200th response from the 12,800 Qwen3-4B mathematics
training trajectories, obtaining 64 responses and 124,923 active
positions.
Response segments are labeled by their function in the surrounding
reasoning using the eight-role ThinkARM taxonomy~\citep{li2026thinkarm}.
Table~\ref{tab:reasoning-role-definitions} summarizes the roles, together
with a Formatting category for structural tokens and whitespace.

\input{tables/reasoning_role_definitions}

Three assistant annotators label 6,648 content segments using the
problem, surrounding response, and a shared codebook, without access to
model scores or construction identities.
An independent assistant reviews 128 sampled segments, with 84.4\%
agreement before adjudication.
The labels are finalized before aggregating signal statistics.
A token receives the label of the segment overlapping the largest
number of its non-whitespace UTF-8 bytes, with ties assigned to the
earlier segment.

\section{Case Studies}
\label{app:case-study}

\subsection{Reasoning Improvements After Self-Distillation}

Figures~\ref{fig:case-study-math} and~\ref{fig:case-study-code} compare
incorrect original-model responses with correct student responses to
the same evaluation prompts, using matched sampling seeds.
The mathematics examples show two changes in reasoning.
Quantization-based training corrects double counting in a symmetric
divisor argument, while pruning-based training yields a formulation that
preserves the distinction between reusing an edge and revisiting a vertex.
The code examples show improvements in algorithm design and constraint
checking.
The students recover the correct deletion-time invariant in
\emph{Gravity} and enforce both initial nonadjacency conditions in
\emph{Switch Seats}.
Each code example includes a recorded test that exposes the original
program's error and is correctly handled by the student's program.

\input{figures/case_study_math}
\input{figures/case_study_code}

\clearpage
\subsection{Local Supervision from Weak-Model Contrasts}

Figure~\ref{fig:case-study-signals} extends the token-level examples in
Figure~\ref{fig:case-study} with two more Qwen3-4B training trajectories.
In the quantization example, the contrast reinforces the negative sign
in an inadmissible prime candidate and the subsequent decision to discard
it.
In the pruning example, it reinforces the zero required by a feasible
variable assignment and the comparison with a lower-valued alternative.
These examples show how the training signal emphasizes intermediate
calculations and the decisions that follow from them.

\input{figures/case_study_signals}

%% file: tables/construction_comparison_full.tex
\begin{table}[!ht]
  \centering
  \caption{Alternative constructions on Qwen3-4B and AIME24 (Avg@32).
  Default RTN and magnitude pruning (shaded) achieve the highest accuracy in
  their families. $g$ denotes group size. Alternatives include
  GPTQ~\citep{frantar2023gptq}, Smooth W4A4~\citep{xiao2023smoothquant},
  Wanda~\citep{sun2024wanda}, and SparseGPT~\citep{frantar2023sparsegpt}.}
  \label{tab:construction-comparison-full}
  \small
  \setlength{\tabcolsep}{4pt}
  \renewcommand{\arraystretch}{1.08}
  \begin{tabular}{>{\raggedright\arraybackslash}p{\dimexpr0.18\textwidth-2\tabcolsep\relax}
      >{\raggedright\arraybackslash}p{\dimexpr0.17\textwidth-2\tabcolsep\relax}
      >{\raggedleft\arraybackslash}p{\dimexpr0.11\textwidth-2\tabcolsep\relax}
      >{\raggedright\arraybackslash}p{\dimexpr0.22\textwidth-2\tabcolsep\relax}
      >{\raggedright\arraybackslash}p{\dimexpr0.21\textwidth-2\tabcolsep\relax}
      >{\raggedleft\arraybackslash}p{\dimexpr0.11\textwidth-2\tabcolsep\relax}}
    \toprule
    \multicolumn{3}{c}{\textit{Quantization}} & \multicolumn{3}{c}{\textit{Pruning}} \\
    \cmidrule(lr){1-3}\cmidrule(lr){4-6}
    Construction & Setting & Acc. (\%) & Construction & Setting & Acc. (\%) \\
    \midrule
    Qwen3-4B & --- & 73.4 & Qwen3-4B & --- & 73.4 \\
    \midrule
    RTN & 6-bit, $g=128$ & 73.0 & Magnitude pruning & 20\% per row & 74.3 \\
    RTN & 4-bit, $g=128$ & 72.0 & Magnitude pruning & 40\% per row & 74.4 \\
    RTN & 4-bit, $g=32$ & 71.0 & Wanda & 20\% per row & 72.3 \\
    GPTQ & 3-bit, $g=128$ & 72.0 & Wanda & 40\% per row & 74.4 \\
    GPTQ & 2-bit, $g=128$ & 70.9 & Wanda & 2:4 & 74.4 \\
    W4A8 & $g=128$ & 72.4 & SparseGPT & 50\% & 73.3 \\
    W4A4 & $g=128$ & 70.0 & SparseGPT & 2:4 & 73.5 \\
    Smooth W4A4 & $g=128$ & 75.3 & MLP channel pruning & 25\% & 73.5 \\
    \rowcolor{methodblue}
    RTN & 3-bit, $g=128$ & \textbf{75.4} & Magnitude pruning & $\tau=1/6$, $g=128$ & \textbf{75.8} \\
    \bottomrule
  \end{tabular}
\end{table}

%% file: tables/row_norm_restoration.tex
\begin{table}[!htbp]
  \centering
  \caption{Effect of row-norm restoration on the pruning construction.
  Accuracy (\%) is measured with Avg@32.
  The default with restoration (shaded) gives a small additional gain
  on both models.}
  \label{tab:row-norm-restoration}
  \small
  \setlength{\tabcolsep}{12pt}
  \renewcommand{\arraystretch}{1.12}
  \begin{tabular}{lcc}
    \toprule
    \multirow{2}{*}{Construction} & Qwen3-4B & Qwen3.5-4B \\
      & AIME25 & HMMT25 \\
    \midrule
    Pruning w/o row-norm restoration & 65.4 & 70.0 \\
    \rowcolor{methodblue}
    Pruning w/ row-norm restoration & \textbf{66.0} & \textbf{70.5} \\
    \bottomrule
  \end{tabular}
\end{table}

%% file: tables/reasoning_role_definitions.tex
\begin{table}[!htbp]
  \centering
  \caption{Contextual roles used for the training-signal analysis.
  The eight reasoning roles follow ThinkARM~\citep{li2026thinkarm}.
  Labels reflect each segment's function in the response.}
  \label{tab:reasoning-role-definitions}
  \small
  \setlength{\tabcolsep}{8pt}
  \begin{tabular}{lp{0.76\textwidth}}
    \toprule
    Role & Function \\
    \midrule
    Read & Restates the given facts, conditions, or requested result. \\
    Analyze & Explains a concept, builds a representation, or makes a logical deduction. \\
    Plan & Commits to an action before carrying it out. \\
    Implement & Executes calculations, substitutions, or other concrete operations. \\
    Explore & Proposes a tentative approach or uncertain possibility. \\
    Verify & Checks an existing result or approach against the problem's conditions. \\
    Monitor & Comments briefly on the reasoning process. \\
    Answer & States a proposed or final answer to the original question. \\
    Formatting & Marks structural tokens or whitespace-only spans. \\
    \bottomrule
  \end{tabular}
\end{table}

%% file: figures/case_study_math.tex
\begin{figure}[!htbp]
  \centering
  \begingroup
  \fontsize{10}{12}\selectfont
  \setlength{\parindent}{0pt}
  \setlength{\parskip}{3pt}
  \setlength{\fboxsep}{3pt}
  \definecolor{caseoriginal}{RGB}{145,83,39}
  \definecolor{casestudent}{RGB}{40,91,135}

  \begin{minipage}{\linewidth}
    \textbf{(a) Counting each symmetric pair once}
    \hfill Qwen3.5-4B $\cdot$ HMMT25

    \textbf{Question.}
    Compute $\displaystyle
      \sum_{j=-1000}^{1000}\left\lfloor\frac{2025}{j+0.5}\right\rfloor$.
    \par\vspace{4pt}

    \begin{minipage}[t]{0.485\linewidth}
      \colorbox[HTML]{F8E8D8}{%
        \parbox{\dimexpr\linewidth-2\fboxsep\relax}{%
          \textcolor{caseoriginal}{\textbf{Original (incorrect)}}}}
      \par\vspace{3pt}
      Thus, there are $15 - 1 = 14$ positive divisors in the range.
      Similarly, there are 14 negative divisors in the range (since
      $-2025 < -1999$).
      Total number of valid $m$ values is $14 + 14 = 28$.
      These correspond to \textbf{28 pairs} where the sum is 0.
      The remaining number of pairs is $1000 - 28 = 972$.
      These pairs sum to -1.
      So the sum over pairs is
      $28 \times 0 + 972 \times (-1) = -972$.

      [\ldots]

      The total sum is:
      \[
        S = -972 + 2 = -970
      \]
      Answer: $\boxed{-970}$
    \end{minipage}
    \hfill
    \begin{minipage}[t]{0.485\linewidth}
      \colorbox{methodblue}{%
        \parbox{\dimexpr\linewidth-2\fboxsep\relax}{%
          \textcolor{casestudent}{%
            \textbf{\method (Quantization, correct)}}}}
      \par\vspace{3pt}
      We restrict our attention to
      $k \in P' = \{1, 3, \dots, 1999\}$.
      The only divisor not in this set is $2025$ itself (since
      $2025 > 1999$).
      All other 14 divisors are $\le 675 < 1999$ and thus included.
      So, there are 14 values of $k$ for which
      $\frac{4050}{k}$ is an integer.
      For these \textbf{14 pairs}, the contribution is $0$.
      The total number of pairs formed this way corresponds to the size
      of $P'$, which is $\frac{1999-1}{2} + 1 = 1000$.
      The number of pairs where the contribution is $-1$ is
      $1000 - 14 = 986$.

      [\ldots]
      \[
        S = -986 + 2 = -984.
      \]
      Answer: $\boxed{-984}$
    \end{minipage}
  \end{minipage}

  \par\vspace{10pt}
  {\color{black!25}\rule{\linewidth}{0.4pt}}
  \par\vspace{6pt}

  \begin{minipage}{\linewidth}
    \textbf{(b) Preserving the edge constraint}
    \hfill Qwen3.5-4B $\cdot$ HMMT25

    \textbf{Question.}
    In an $11\times11$ grid, adjacent cells share a door.
    Find the maximum number of doors in a closed walk that never reuses a door.
    \par\vspace{4pt}

    \begin{minipage}[t]{0.485\linewidth}
      \colorbox[HTML]{F8E8D8}{%
        \parbox{\dimexpr\linewidth-2\fboxsep\relax}{%
          \textcolor{caseoriginal}{\textbf{Original (incorrect)}}}}
      \par\vspace{3pt}
      The problem asks for the maximum number of doors Karthik can walk
      through in an $11 \times 11$ grid, starting and ending at the same
      cell, without reusing any door.
      This is equivalent to finding the length of the \textbf{longest
      cycle} in the grid graph $G_{11,11}$.

      [\ldots]

      The graph is bipartite, which implies that any cycle must have
      an even length.
      Furthermore, a cycle of length $L$ must visit $L/2$ vertices from
      set $A$ and $L/2$ vertices from set $B$.
      Therefore, the length of any cycle $L$ satisfies
      $L/2 \le |B| = 60$, which implies $L \le 120$.

      [\ldots]

      Answer: $\boxed{120}$
    \end{minipage}
    \hfill
    \begin{minipage}[t]{0.485\linewidth}
      \colorbox{methodblue}{%
        \parbox{\dimexpr\linewidth-2\fboxsep\relax}{%
          \textcolor{casestudent}{%
            \textbf{\method (Pruning, correct)}}}}
      \par\vspace{3pt}
      The problem asks for the maximum number of doors (edges) Karthik
      can traverse in an \textbf{Eulerian circuit} (a closed walk visiting
      every traversed door exactly once) in an $11 \times 11$ grid graph.
      This is equivalent to finding a subgraph $H$ of the grid graph
      $G$ that is connected and has all vertices of even degree,
      maximizing the number of edges $|E(H)|$.

      [\ldots]

      Total number of removed edges
      $|F| = 16 (\text{internal}) + 4 (\text{cross}) = 20$.

      [\ldots]

      Total number of doors (edges) in an $11 \times 11$ grid is
      $2 \times 11 \times (11-1) = 220$.
      Maximum doors Karthik can walk = Total edges - Minimum removed edges
      \[
        220 - 20 = 200.
      \]
      [\ldots]

      Answer: $\boxed{200}$
    \end{minipage}
  \end{minipage}
  \endgroup
  \caption{\textbf{Reasoning improvements after self-distillation.}
  On HMMT25, the quantization-trained Qwen3.5-4B student counts each
  symmetric divisor pair once, correcting the original model's
  double counting (a).
  The pruning-trained student models the door constraint as an
  Eulerian circuit, allowing repeated visits to cells while traversing
  each selected door once (b).
  Excerpts are taken from the original and trained models' responses
  to the same questions, with omissions marked by [\ldots].}
  \label{fig:case-study-math}
\end{figure}

%% file: figures/case_study_code.tex
\begin{figure}[!t]
  \centering
  \begingroup
  \setlength{\fboxsep}{5pt}
  \setlength{\tabcolsep}{0pt}
  \newcommand{\codecasehead}[2]{%
    \colorbox[HTML]{#1}{%
      \makebox[\dimexpr\linewidth-2\fboxsep\relax][l]{%
        \normalfont\small\bfseries #2}}}
  \newcommand{\codeline}[2]{%
    \hangindent=\dimexpr #1em+1.2em\relax\hangafter=1
    \hspace*{#1em}\texttt{\detokenize{#2}}\\}
  \newcommand{\codefocusline}[2]{%
    \hangindent=\dimexpr #1em+1.2em\relax\hangafter=1
    \hspace*{#1em}\texttt{\bfseries\detokenize{#2}}\\}

  \begin{minipage}{\linewidth}
    \small
    \textbf{(a) Recovering the deletion-time invariant.}
    \hfill Qwen3.6-35B-A3B, quantization
    \par\smallskip
    \textbf{Problem.}
    At each time step, remove a full bottom row, then process blocks from
    bottom to top, moving each down one cell if the cell below is empty.
    Determine whether a queried block still exists.
    \par\smallskip
    \begin{minipage}[t]{0.485\linewidth}
      \codecasehead{F8E8D8}{Original (incorrect)}
      \par\smallskip
      \fontsize{9}{11.5}\selectfont
      \begin{tabular}{@{}>{\raggedright\arraybackslash}p{\linewidth}@{}}
        \codeline{0}{for m in range(1, K_max + 1):}
        \codeline{1.2}{max_gap = 0}
        \codeline{1.2}{for g_list in gaps:}
        \codeline{2.4}{gap = g_list[m-1]}
        \codeline{2.4}{if gap > max_gap:}
        \codeline{3.6}{max_gap = gap}
        \codefocusline{1.2}{current_T += max_gap}
        \codeline{1.2}{T[m] = current_T}
      \end{tabular}
    \end{minipage}\hfill
    \begin{minipage}[t]{0.485\linewidth}
      \codecasehead{DDEBF7}{\method (correct)}
      \par\smallskip
      \fontsize{9}{11.5}\selectfont
      \begin{tabular}{@{}>{\raggedright\arraybackslash}p{\linewidth}@{}}
        \codeline{0}{removal_times = list(col_ys[0][:M])}
        \codeline{0}{for ys in col_ys[1:]:}
        \codeline{1.2}{for r in range(M):}
        \codefocusline{2.4}{if ys[r] > removal_times[r]:}
        \codefocusline{3.6}{removal_times[r] = ys[r]}
      \end{tabular}
    \end{minipage}
    \par\smallskip
    \textbf{Failing test.}
    Column heights are $(31,39,43,48,50)$ and $(2,35,39,44)$.
    For the second block in column~2, the original predicts deletion at
    $31+\max(8,33)=64$, while the student obtains $\max(39,35)=39$.
    At time $48.5$, the answers are therefore
    \textbf{Yes} (incorrect) and \textbf{No} (correct).
  \end{minipage}

  \par\vspace{10pt}
  {\color[gray]{0.75}\hrule}
  \vspace{10pt}

  \begin{minipage}{\linewidth}
    \small
    \textbf{(b) Enforcing both initial conditions.}
    \hfill Qwen3.5-4B, pruning
    \par\smallskip
    \textbf{Problem.}
    Each couple appears twice in a seating array.
    Count pairs of couples whose members are initially nonadjacent
    and can each become adjacent by swapping their four seats.
    \par\smallskip
    \begin{minipage}[t]{0.485\linewidth}
      \codecasehead{F8E8D8}{Original (incorrect)}
      \par\smallskip
      \fontsize{9}{11.5}\selectfont
      \begin{tabular}{@{}>{\raggedright\arraybackslash}p{\linewidth}@{}}
        \codeline{0}{for i in range(2 * N - 1):}
        \codeline{1.2}{u = A[i]}
        \codeline{1.2}{v = A[i+1]}
        \codefocusline{1.2}{if u != v:}
        \codeline{2.4}{[...]}
        \codeline{2.4}{groups[key].append(i)}
        \codeline{0}{[...]}
        \codeline{0}{for j in range(n_indices):}
        \codeline{1.2}{[...]}
        \codeline{1.2}{if k >= 0:}
        \codeline{2.4}{current_count += (k + 1)}
        \codeline{0}{ans += current_count}
      \end{tabular}
    \end{minipage}\hfill
    \begin{minipage}[t]{0.485\linewidth}
      \codecasehead{DDEBF7}{\method (correct)}
      \par\smallskip
      \fontsize{9}{11.5}\selectfont
      \begin{tabular}{@{}>{\raggedright\arraybackslash}p{\linewidth}@{}}
        \codeline{0}{for a in range(1, N + 1):}
        \codefocusline{1.2}{if pos[a][1] - pos[a][0] <= 1:}
        \codeline{2.4}{continue}
        \codeline{0}{[...]}
        \codeline{0}{if abs(other_pos_b - p2) == 1:}
        \codeline{1.2}{if b_val not in local_seen:}
        \codeline{2.4}{local_seen.add(b_val)}
        \codeline{0}{[...]}
        \codefocusline{0}{if pos[b_val][1] - pos[b_val][0] > 1 and b_val > a:}
        \codeline{1.2}{ans += 1}
      \end{tabular}
    \end{minipage}
    \par\smallskip
    \textbf{Failing test.}
    For $A=[2,1,1,2]$, the original counts the two disjoint mixed-couple
    edges and returns \textbf{1}.
    Couple~1 is already adjacent, so this pair is invalid.
    The student checks both couples and returns the correct answer,
    \textbf{0}.
  \end{minipage}
  \endgroup
  \caption{\textbf{Code excerpts before and after self-distillation on
  LiveCodeBench v6.}
  On \emph{Gravity} (a), quantization-based training replaces an incorrect
  accumulation of inter-block gaps with the maximum initial height at each
  rank.
  On \emph{Switch Seats} (b), pruning-based training preserves both
  nonadjacency conditions when translating the problem into code.
  Both post-training programs pass the corresponding test suites.
  The examples illustrate improved algorithmic reasoning and constraint
  checking.}
  \label{fig:case-study-code}
\end{figure}

%% file: figures/case_study_signals.tex
\begin{figure}[!htbp]
  \centering
  \includegraphics[width=\textwidth]{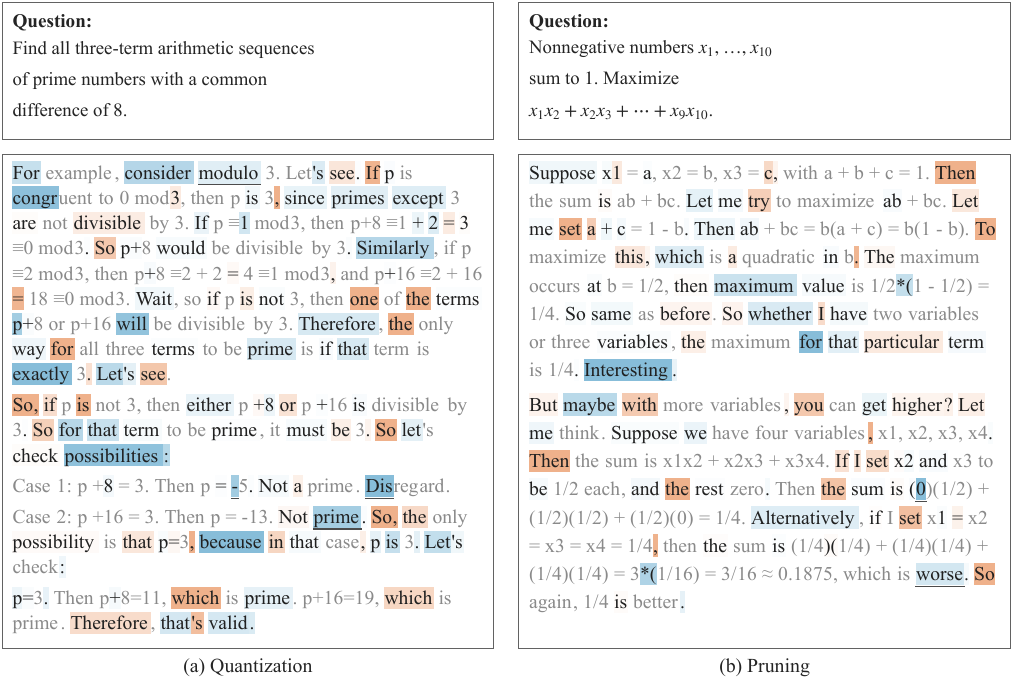}
  \caption{\textbf{Weak-model contrasts reinforce local reasoning steps.}
  Blue and orange indicate increased and decreased target probability
  relative to the behavior distribution, with underlines marking the
  tokens discussed in the text.
  Quantization reinforces the negative sign in $p=-5$ and the rejection
  of the resulting nonprime candidates (a).
  Pruning reinforces the zero used when substituting
  $x_2=x_3=1/2$ and all other variables equal to zero, together with
  the comparison identifying a uniform four-variable allocation as
  worse (b).
  The highlighted steps connect numerical calculations to candidate
  rejection and comparisons between feasible assignments.}
  \label{fig:case-study-signals}
\end{figure}